\documentclass[times, review, 10pt]{elsarticle}

\usepackage{amssymb}
\usepackage{amsmath}

\usepackage{algpseudocode}
\usepackage{algorithm}
\usepackage{booktabs}
\usepackage{multirow}
\usepackage{makecell}
\usepackage{xcolor}
\usepackage{caption}
\usepackage[hidelinks]{hyperref}

\definecolor{text-red}{RGB}{255, 0, 0}
\definecolor{text-blue}{RGB}{0, 0, 255}

\newcommand{\eqautoref}[1]{\hyperref[#1]{Eq.~(\ref*{#1})}}

\newcommand{\best}[1]{\textbf{#1}}
\newcommand{\second}[1]{\underline{#1}}
\newcommand{\lossmark}[1]{\underline{#1}}
\newcommand{\vpara}[1]{\vspace{0.05in}\noindent\textbf{#1}}

\def\model{\textbf{M2Patch}}

\begin{document}

\begin{frontmatter}
    
    \title{Structured Latent Space Modeling over Multi-Scale Temporal Patches for Multivariate Time Series Forecasting}

    \author[1]{Xingsheng Chen}
    \author[2]{Deyu Yi}
    \author[1]{Siu-Ming Yiu}

    \affiliation[1]{organization={School of Computing and Data Science, The University of Hong Kong},
                    city={Hong Kong SAR},
                    country={China}}

    \affiliation[2]{organization={Innovation Engineering College, Macau University of Science and Technology},
                    city={Macau},
                    country={China}}

    \begin{abstract}
        Existing patching and multi-scale methods advance multivariate time series forecasting but treat learned representations as transient byproducts of prediction, lacking explicit mechanisms that enforce structural consistency across temporal scales. We propose \model, a CNN-based architecture that organizes channel-independent observations into a structured latent space via two complementary differentiable penalties. Multi-scale patching decomposes the input into overlapping temporal granularities, depthwise separable CNN blocks with progressively growing dilation extracts scale-specific features at linear complexity, and per-scale learned projections compress these features into a compact latent representation. An intra-scale smoothness penalty enforces temporal continuity between adjacent patches, while an inter-scale alignment penalty restores cross-granularity interaction through learnable cross-scale mappings, so that all scales encode mutually consistent representations of the underlying dynamics. Extensive experiments on ten real-world benchmark datasets demonstrate that \model\ significantly outperforms state-of-the-art baselines. Further analyses establish \model\ as a structure-aware recognizer: it recovers channel functional groupings and remains robust under patch-level input corruption, confirming that the structured latent space captures the data's intrinsic dynamics. The code is available at~\url{https://github.com/XsChen524/m2patch}.
    \end{abstract}

    \begin{keyword}
    Multivariate Time Series Forecasting \sep Latent Space Modeling \sep Cross-Scale Alignment \sep Channel-Independent Models \sep Representation Learning
    \end{keyword}

\end{frontmatter}

\section{Introduction}

\begin{figure*}[!t]
    \centering
    \includegraphics[width=\textwidth]{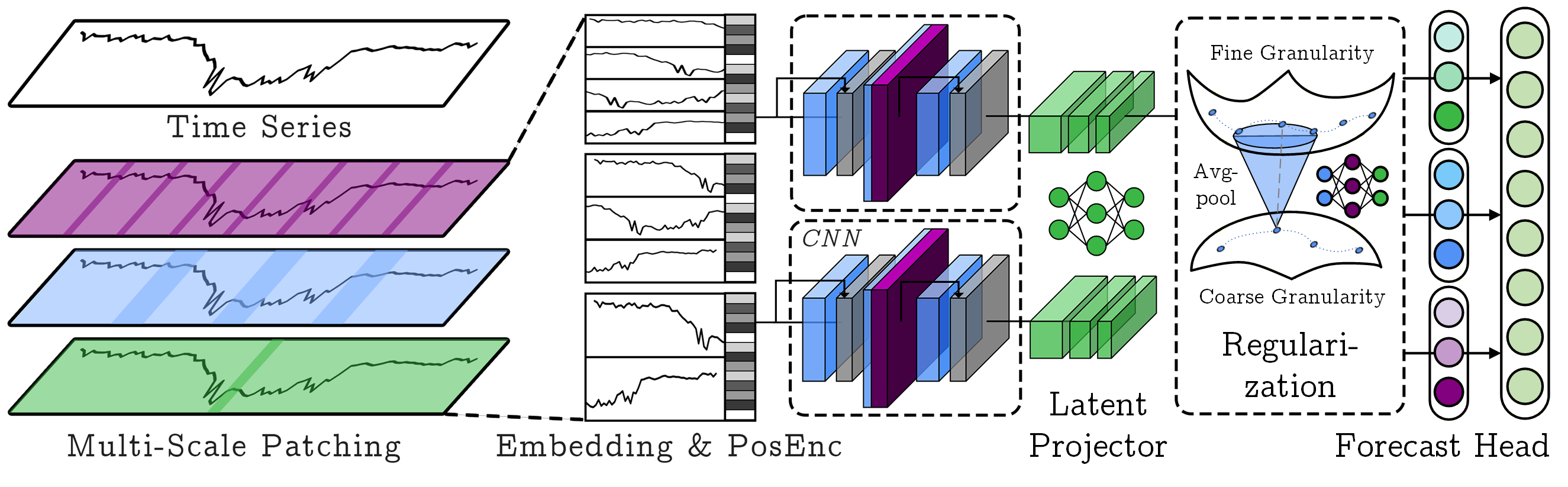}
    \caption{Illustration of \model's key designs: multi-scale patch decomposition, the depthwise separable CNN backbone, cross-scale alignment and latent-space regularization, and multi-scale forecast fusion.}
    \label{fig:teaser}
\end{figure*}

Multivariate time series forecasting underpins decision-making in a wide range of real-world domains, including energy management, traffic flow control, financial risk assessment, and weather prediction, where anticipating future trends is essential. Learning such forecasts requires representations that capture structural patterns and dependency relationships across multiple temporal granularities. In complex dynamical systems, whether physical, biological, or industrial, observations capture the joint evolution of multiple variables from rapid fluctuations to long-term trends. A principled data-driven architecture must therefore (i)~decompose and understand signals from different time scales; (ii)~encode intrinsic dynamic features into structured latent representations; and (iii)~enforce structural consistency across scales, such that it produces representations that support accurate forecasting and exhibit interpretable latent structure simultaneously, as illustrated in Fig.~\ref{fig:teaser}.

Recent work on time series forecasting has progressed along a trajectory from Transformer-based architectures toward increasingly efficient and structurally-aware designs. Standard Transformer methods rely on self-attention for global dependency modeling but incur $\mathcal{O}(L^2)$ complexity: Autoformer~\cite{wu2021autoformer} substitutes period-aware auto-correlation for point-wise attention, iTransformer~\cite{liu2023itransformer} inverts attention to operate across variates, and Crossformer~\cite{zhang2023crossformer} applies two-stage attention over both the temporal and variable dimensions. Patching approaches~\cite{nie2023patchtst} address this bottleneck by segmenting inputs into subseries-level tokens, substantially reducing the effective sequence length. Multi-scale architectures route information across scales in complementary ways: TimeMixer~\cite{wang2024timemixer} cascades decomposable MLPs across temporal resolutions, and MRN~\cite{guo2025mrn} supplies coarse-to-fine recurrent guidance across scales. Meanwhile, state-space models~\cite{gu2023mamba} offer linear-time alternatives at the cost of specialized hardware-aware implementations. Despite these advances, three challenges remain for multi-scale forecasting. First, attention-based backbones retain quadratic cost in the sequence length, and existing patching schemes operate at a single fixed granularity without explicit cross-scale interaction. Second, multi-scale and frequency-aware methods exchange information across scales through hidden-state mixing inside the backbone, without isolating a separately regularizable latent family whose cross-scale alignment is explicitly constrained. Third, learned intermediate representations are treated as transient byproducts of prediction, and the structural information they encode about temporal patterns and cross-variable dependencies is neither constrained nor retained for downstream analysis.

To address these challenges, we propose Multi-scale Patching (\model), a patching-based forecasting architecture built on three design principles. First, multi-scale patch-wise decomposition partitions the input into $K$ temporal scales with distinct patch sizes and strides, capturing patterns from fine-grained fluctuations to coarse-grained trends within a unified framework. Second, a depthwise separable CNN backbone with exponentially growing dilation replaces self-attention at each temporal scale independently, enabling scale-specific feature extraction while achieving linear complexity, on par with other CNN-based patching methods. Third, the latent space is organized by two complementary auxiliary regularization terms: an intra-scale smoothness term that encourages gradual transitions between temporally adjacent latent patches, and an inter-scale consistency term that enforces structural alignment between fine-grained and coarse-grained representations through learned cross-scale mappings. These mappings let the channel-independent representations at different temporal scales interact directly, complementing the robustness of channel independence (CI) with low-overhead cross-scale information exchange. These terms model the intrinsic relationships among patches within a compact latent space, adapting to each dataset's underlying dynamics.

Our primary contributions are summarized as follows.
\begin{enumerate}
    \item We formalize a \emph{structured multi-scale latent space} via two structural properties, intra-scale temporal smoothness and inter-scale alignment, and realize it with a channel-independent CNN that enforces these properties through differentiable $\ell_2$ surrogates and learned fine-to-coarse correspondence maps.
    \item We design \model, a CNN-based architecture that realizes this framework via multi-scale patching, depthwise separable convolutions with progressive dilation, per-scale latent projection heads, and lightweight cross-scale mappings that enable interaction across temporal scales under the channel-independent design.
    \item We show that the auxiliary constraints induce a compact latent space that explicitly models the intrinsic relationships across patches and scales, encoding the data's latent dynamics to regularize multi-scale patching for forecasting with competitive accuracy and stability.
    \item We conduct experiments on ten real-world benchmarks spanning diverse domains and variable counts, showing that \model\ achieves competitive or superior accuracy against representative patching and Transformer baselines, and validating its design through efficiency, ablation, robustness, and multi-scale case studies.
\end{enumerate}

The rest of this paper is organized as follows. Section~\ref{sec:related_work} reviews related work, and Section~\ref{sec:preliminary} formulates the multivariate forecasting problem. Section~\ref{sec:methodology} presents the proposed \model\ architecture and the structured latent space formulation. Section~\ref{sec:experiments} reports comparative results, ablation studies, and other relevant analyses. Section~\ref{sec:conclusion} concludes this paper.

\section{Related work}
\label{sec:related_work}

\subsection{Transformer-based forecasting}

Classical statistical methods such as ARIMA~\cite{shumway2017arima} and VAR~\cite{zivot2006vector} provided foundational tools for time series analysis, but their linear assumptions limit the capacity to capture complex nonlinear temporal dependencies. The Transformer architecture~\cite{vaswani2017attention} and its forecasting-oriented variants, surveyed in~\cite{wen2023transformers}, advanced long-horizon accuracy: Informer~\cite{zhou2021informer} introduces ProbSparse self-attention to sparsify the attention matrix, Autoformer~\cite{wu2021autoformer} replaces point-wise attention with an auto-correlation mechanism over delayed phases, Crossformer~\cite{zhang2023crossformer} applies two-stage attention across both the temporal and variable dimensions, and iTransformer~\cite{liu2023itransformer} inverts the conventional design by tokenizing each variate and attending across the spatial dimension. Beyond attention variants, decomposition-based fusion networks~\cite{zhang2023dfnet} integrate explicit series decomposition into dense fusion architectures for long-sequence forecasting. Despite these advances, the quadratic complexity of self-attention in the sequence length remains a bottleneck for long-horizon forecasting.

\subsection{Efficiency-oriented architectures}

This cost has motivated a family of efficiency-oriented architectures. DLinear~\cite{zeng2023transformers} shows that simple trend-seasonal decomposition with single-layer linear networks can match Transformer-level accuracy, and TiDE~\cite{das2023long} attains comparable performance through dense MLP encoder-decoder stacks, together questioning the necessity of costly attention. Convolutional alternatives preserve local temporal detail at full resolution: non-pooling temporal convolutional architectures~\cite{liu2019nonpooling} retain fine-grained activations, modernized pure-convolution structures~\cite{donghao2024moderntcn} enlarge the effective receptive field, and dilated convolutional networks~\cite{zhang2023ctfnet} integrate multi-scale dilated convolution to capture both global and local context without resolution loss. State space models~\cite{gu2023mamba} offer a further linear-time alternative through selective scanning, and their forecasting effectiveness has been validated by recent empirical studies~\cite{wang2025mamba}. These developments motivate the search for architectures that combine computational efficiency with explicit structural priors, a direction pursued by patching-based approaches that reduce the effective sequence length before feature extraction.

\subsection{Multi-scale and channel independence}

Patching-based methods restructure time series into subseries-level tokens, substantially reducing the sequence length processed by the backbone. PatchTST~\cite{nie2023patchtst} introduced this approach by segmenting univariate sequences into fixed-length patches and processing them channel-independently through a standard Transformer encoder, achieving strong forecasting accuracy while partially mitigating the $\mathcal{O}(L^2)$ attention cost. TimesNet~\cite{wu2023timesnet} captures multi-periodicity through TimesBlock, a parameter-efficient Inception block that applies 2D convolutional kernels to capture both intra- and inter-period variations within 2D tensors organized by Fourier-detected dominant periods. In parallel, patching variants have diversified toward periodic dynamics modeling~\cite{tan2025pdfnet}, frequency-aware patch embeddings~\cite{peng2026periodpatch}, codebook-assisted patch quantization~\cite{ma2026recast}, and multi-resolution patch-based MLP fusion~\cite{bi2026patchfusionmlp}. A common limitation, however, is that most patching methods operate at a single fixed granularity and neglect the complementary information carried by coarser temporal scales.

TimeMixer~\cite{wang2024timemixer} extends the multi-scale perspective through decomposable MLP-based mixing across temporal resolutions, and subsequent work pursues multi-scale decomposition along complementary axes, including wavelet-based attention~\cite{wei2025waveformer}, multiresolution Transformers~\cite{zhu2025mrtransformer}, coarse-to-fine cross-scale recurrent networks~\cite{guo2025mrn}, and multi-scale time-variable interaction~\cite{liu2025mstvi}. Frequency-aware designs further strengthen long-range dependency modeling through frequency-enhanced decomposition~\cite{zhou2022fedformer} and time-frequency decomposition~\cite{luo2025tfdnet}. Yet these methods exchange scale information implicitly through hidden-state mixing inside the backbone, so the cross-scale interaction remains entangled with the prediction pathway and yields no intermediate representation that could be regularized or analyzed independently.

A notable trend across these designs is the adoption of channel independence: high-performing models such as PatchTST and DLinear process each variable independently, motivated by robustness to distribution shift (achieved by disentangling shift effects~\cite{fan2023dishts} or reversible instance normalization~\cite{kim2022reversible}) and by reduced overfitting. By construction, however, CI discards inter-variable interactions, which is suboptimal for datasets with strong cross-variable dependencies: dynamic spatio-temporal graph transformers~\cite{han2025dygraphformer} couple cross-variable dependencies with temporal dynamics, fully-connected spatio-temporal graphs~\cite{wang2026gaussian} explicitly capture correlations between different sensors at different times, DisenTS~\cite{liu2026disents} restores such interactions by disentangling evolving channel patterns, and the joint spatiotemporal framework of~\cite{yuan2022single} learns cross-variable features together with temporal dynamics. Recovering such dependencies therefore calls for post-hoc analysis of learned latent representations rather than architectural coupling.

\subsection{Representation learning}

Beyond forecasting accuracy, learned time series representations are increasingly evaluated for intrinsic structural properties such as clustering separability, transferability, and robustness to perturbations. Self-supervised methods impose consistency and contrastive objectives that shape the latent organization for downstream use: TS2Vec~\cite{yue2022ts2vec} learns universal representations through temporal and instance-wise contrastive losses, multi-scale self-supervised learning with temporal alignment~\cite{chen2024mssrl} shapes latent organization across temporal scales, topological contrastive learning~\cite{kim2026topocl} preserves temporal semantics under augmentation, and GNN-enhanced contrastive learning~\cite{wang2024contrastive} captures inter-variate structure for universal representations. \model\ adopts a complementary, regularization-based approach: rather than enforcing consistency across augmented views, it enforces temporal smoothness and cross-scale alignment within a single forward pass, producing a latent organization whose properties we empirically examine in Section~\ref{sec:clustering}. Compared with coarse-to-fine recurrent guidance~\cite{guo2025mrn} and MLP-based scale mixing~\cite{wang2024timemixer}, \model\ differs along three axes: it propagates information from fine to coarse scales, enforces cross-scale alignment through an explicit differentiable $\ell_2$ penalty rather than hidden-state mixing, and retains a channel-independent CNN backbone.

\section{Preliminary}
\label{sec:preliminary}

\subsection{Problem Formulation}

Let $\mathbf{X} = [\mathbf{x}_1, \mathbf{x}_2, \ldots, \mathbf{x}_L] \in \mathbb{R}^{L \times N}$ denote a multivariate time series of $L$ historical time steps with $N$ variables, where $\mathbf{x}_t = [x_t^{1}, x_t^{2}, \ldots, x_t^{N}] \in \mathbb{R}^{N}$ collects the observations of $N$ correlated signals at time $t$. Given the $L$-step lookback window, the forecasting objective is to predict the next $P$ time steps:
\begin{equation}
    \hat{\mathbf{Y}} = [\hat{\mathbf{x}}_{L+1}, \hat{\mathbf{x}}_{L+2}, \ldots, \hat{\mathbf{x}}_{L+P}] = f_\theta(\mathbf{X}_{1:L}) \in \mathbb{R}^{P \times N}
\end{equation}
where $f_\theta(\cdot)$ is a learnable forecasting model parameterized by $\theta$. Throughout this paper, we view $f_\theta$ as inducing an intermediate representation of $\mathbf{X}$ whose structure encodes the discovered temporal patterns and inter-variable dependencies, so that accurate forecasting and interpretable pattern recognition become joint objectives of a single learned representation.

\section{Methodology}
\label{sec:methodology}

\begin{figure*}[!t]
    \centering
    \includegraphics[width=\textwidth]{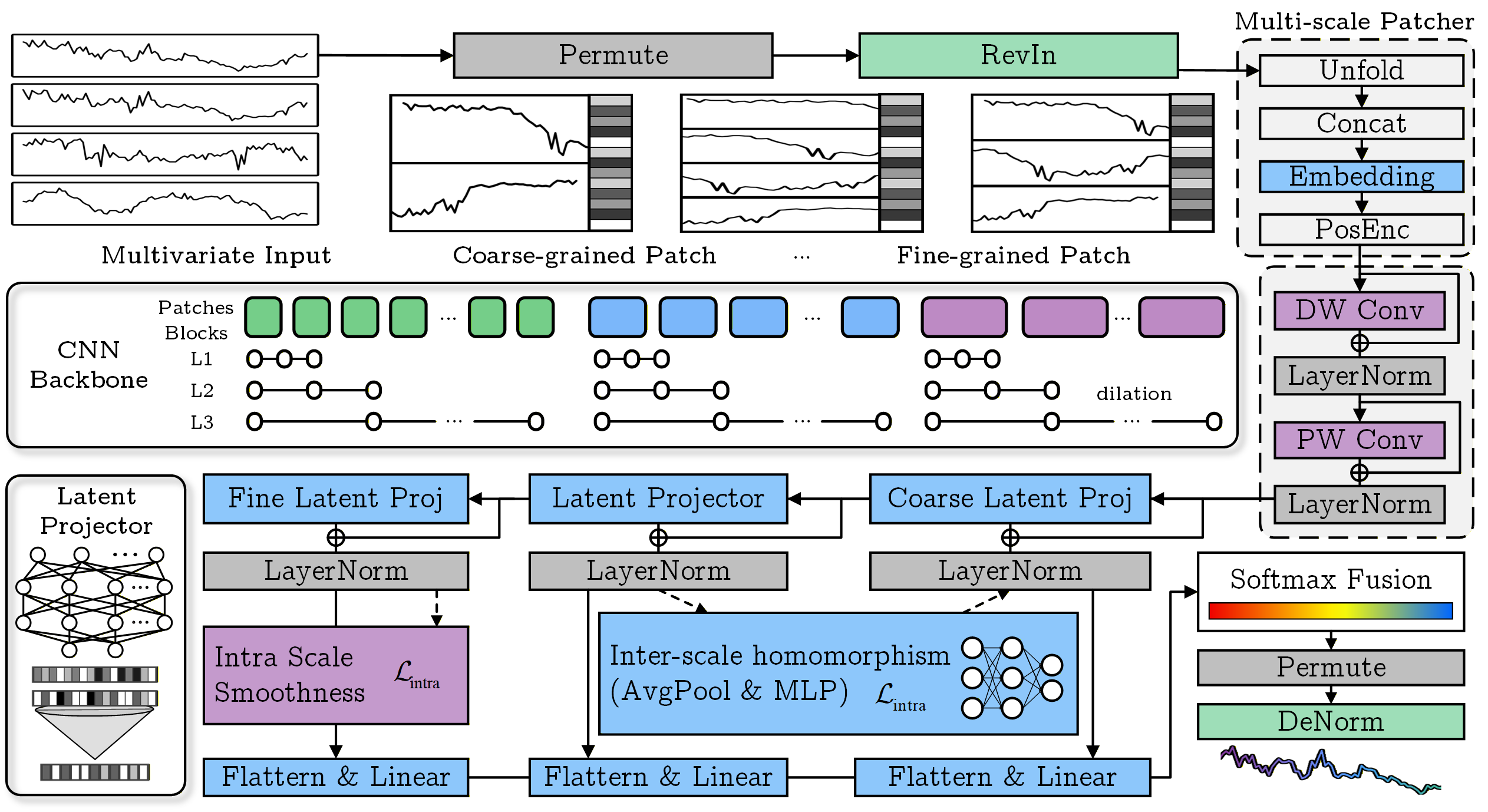}
    \caption{The overall architecture of \model, a forecasting model that organizes multi-scale temporal patches into a structured latent space.}
    \label{fig:framework}
\end{figure*}

The central difficulty stems from \emph{multi-scale coupling}: different variables evolve with distinct characteristic time constants, where rapid transients and slow trends coexist within the same observation window, and a single-scale representation cannot simultaneously resolve both regimes. \model\ addresses this by partitioning the input into a family of overlapping temporal patches with complementary lengths and strides, where fine-grained patches encode high-frequency local transients and coarse-grained patches encode low-frequency trend envelopes. The full pipeline is illustrated in Fig.~\ref{fig:framework}.

At each scale, a depthwise separable CNN backbone (Section~\ref{sec:cnn_backbone}) extracts scale-specific features, which are projected to a compact latent space via learned nonlinear mappings. We organize this latent space through two complementary structural constraints: \emph{temporal continuity}, which maps temporally adjacent patches to nearby latent points, and \emph{cross-scale alignment}, which aligns fine-scale representations with their coarse-scale counterparts; the resulting representation preserves multi-scale temporal structure while remaining compact. The formulation operates entirely in standard Euclidean space, allowing latent organization to emerge autonomously from data-driven constraints.

\subsection{Multi-Scale Temporal Decomposition}
\label{sec:multi_scale}

The input $\mathbf{X} \in \mathbb{R}^{L \times N}$ is first normalized via reversible instance normalization~\cite{kim2022reversible} to mitigate non-stationary distribution shifts that would otherwise perturb the regularization framework. We then decompose the normalized sequence into $K$ temporal scales through a family of sliding-window sampling operators.

For scale $s$, let $P_s$ denote the patch length and $S_s$ the stride between consecutive patch starting positions. The patch decomposition operator $\Pi_s$ extracts a sequence of overlapping temporal segments:
\begin{equation}
    \Pi_s(\mathbf{X}) = \big\{ \mathbf{X}[\,t : t+P_s\,] \;\big|\; t \in \mathcal{T}_s \big\} \in \mathbb{R}^{(B \times N) \times N_s \times P_s}
\end{equation}
where $\mathcal{T}_s = \{0, S_s, 2S_s, \dots, (N_s-1)\cdot S_s\}$ is the set of sampling offsets and $N_s$ the resulting number of patches.

The multi-scale patch family $\{\Pi_s(\mathbf{X})\}_{s=1}^{K}$ provides a complementary set of temporal representations: patches with small $P_s$ encode high-frequency local transients, while patches with large $P_s$ compactly represent low-frequency trend envelopes. The condition $S_s < P_s$ introduces temporal overlap between adjacent patches within each scale, which ensures that the intra-scale smoothness constraint is meaningful: adjacent patches sharing observational context ought to map to nearby points in the latent space.

Patches from all $N$ variables are processed jointly under shared convolutional parameters, following a channel-independent design~\cite{nie2023patchtst} that isolates per-variable temporal structure from cross-variable interactions; encoding each variable's temporal evolution into an independent latent trajectory preserves variable-level interpretability. Each scale then applies an independent linear embedding $\mathbf{W}_{\text{emb}}^s \in \mathbb{R}^{P_s \times d}$ that lifts patches to a common $d$-dimensional representation space, augmented by a learnable positional encoding $\mathbf{W}_{\text{pos}}^s$:
\begin{equation}
    \mathbf{H}_s = \sigma\big(\Pi_s(\mathbf{X}) \cdot \mathbf{W}_{\text{emb}}^s + \mathbf{W}_{\text{pos}}^s\big) \in \mathbb{R}^{(B \times N) \times N_s \times d}
\end{equation}

\subsection{Depthwise Separable Temporal Convolutions}
\label{sec:cnn_backbone}

Each scale's embedded patches are processed by an independent CNN backbone composed of $E$ stacked convolutional blocks with exponentially growing dilation. Within each block, two factorized transformations refine the representation:

\vpara{Depthwise temporal convolution.} A one-dimensional convolution with $d$ channel-wise groups processes each feature channel independently:
\begin{equation}
    \mathbf{H}' = \text{Norm}\big(\mathbf{H} + \text{Conv1d}_{d\text{-group}}(\mathbf{H};\, k,\, \delta)\big)
\end{equation}
where $k$ is the kernel size and $\delta$ the dilation factor. The group-wise design ensures that temporal mixing operates within each channel separately, leaving cross-channel interactions to the subsequent pointwise stage.

\vpara{Pointwise feed-forward transform.} A two-layer channel-mixing sub-network with nonlinear activation $\sigma$ projects features across channels:
\begin{equation}
    \mathbf{H}'' = \text{Norm}\big(\mathbf{H}' + \mathbf{W}_2 \cdot \sigma(\mathbf{W}_1 \cdot \mathbf{H}')\big)
\end{equation}

The dilation grows as $\delta_\ell = 2^{\ell-1}$, so shallow blocks capture fine-grained correlations while deep blocks integrate long-range context at linear cost in the sequence length (Section~\ref{sec:complexity}); the depthwise-pointwise split further separates temporal aggregation from channel mixing, keeping the backbone lightweight.

\subsection{Structured Latent Space Modeling via Auxiliary Constraints}
\label{sec:latent_regularization}

The preceding stages produce $K$ scale-specific representations $\{\mathbf{H}_s''\}$ in a high-dimensional space. The core contribution of \model\ is a framework that organizes this latent space through learned projections and differentiable structural penalties.

\noindent Let $\mathbf{z}^{(s)} = [\mathbf{z}^{(s)}_1, \ldots, \mathbf{z}^{(s)}_{N_s}] \in \mathbb{R}^{N_s \times d_l}$ denote the latent representation at scale $s \in \{1,\ldots,K\}$, and $\mathcal{Z} = \{\mathbf{z}^{(s)}\}_{s=1}^{K}$. We organize $\mathcal{Z}$ into a structured latent representation by requiring two structural properties.

\vpara{Intra-scale temporal smoothness.} Adjacent patches that share observational overlap map to nearby latent points, imposing temporal continuity on the latent trajectory:
\begin{equation}
    E_{\text{intra}}(\mathbf{z}^{(s)}) := \frac{1}{N_s-1} \sum_{t=1}^{N_s-1} \big\| \mathbf{z}^{(s)}_t - \mathbf{z}^{(s)}_{t+1} \big\|_2^2
\end{equation}

\vpara{Inter-scale alignment.} There exist correspondence maps $\Phi = \{\Phi_s\}_{s=1}^{K-1}$ under which fine-scale representations can be aligned with their coarse-scale counterparts: a coarse-scale representation is approximated by the temporally aggregated fine-scale representation under a learned correspondence map:
\begin{equation}
    E_{\text{inter}}(\mathbf{z}^{(s)}, \mathbf{z}^{(s+1)}; \Phi_s) := \frac{1}{N_{s+1}} \sum_{t=1}^{N_{s+1}} \big\| \Phi_s\!\big(\text{Pool}(\mathbf{z}^{(s)})\big)_t - \mathbf{z}^{(s+1)}_t \big\|_2^2
\end{equation}

\noindent These properties define a target latent organization decoupled from any algorithm, realized by minimizing two differentiable $\ell_2$ surrogates, defined below, jointly with the forecasting objective. The surrogates are architecture-agnostic, applying to any multi-scale backbone that produces per-scale latents, and optimization proceeds entirely in standard Euclidean space.

\subsubsection{Latent Space Projection}

For each scale $s$, the CNN backbone output $\mathbf{H}_s'' \in \mathbb{R}^{N_s \times d}$ is mapped to a compact $d_l$-dimensional latent space via a learned projection $\varphi_s$:
\begin{equation}
    \mathbf{z}^{(s)} = \varphi_s(\mathbf{H}_s'') = \text{Norm}\big(f_s(\mathbf{H}_s'') + g_s(\mathbf{H}_s'')\big) \in \mathbb{R}^{N_s \times d_l}
\end{equation}
where $f_s: \mathbb{R}^d \to \mathbb{R}^{d_l}$ is a two-layer nonlinear mapping and $g_s: \mathbb{R}^d \to \mathbb{R}^{d_l}$ is a linear residual projection that preserves a direct gradient pathway through the compression bottleneck. The normalization step stabilizes the latent features for subsequent regularization.

The projection from $d$ to $d_l < d$ performs \emph{dimensionality compression}: the model compresses features while retaining structural information most discriminative for forecasting. Each scale employs an independent projection head $\varphi_s$, reflecting the observation that different temporal granularities produce qualitatively distinct feature distributions: fine-scale representations encode rapid fluctuations, while coarse-scale representations capture smoother trend dynamics. A shared projection would conflate these distinct representational characteristics, suppressing scale-specific information. The cross-scale mappings act as bridges between scale-specific latent spaces, and the consistency terms enforce their alignment through compact two-layer MLPs.

\subsubsection{Temporal Continuity Constraint}

The intra-scale smoothness term enforces temporal continuity within each scale's latent representation, averaging the per-scale energy $E_{\text{intra}}$ across all $K$ scales: $\mathcal{L}_{\text{intra}} = \frac{1}{K} \sum_{s=1}^{K} E_{\text{intra}}(\mathbf{z}^{(s)})$.

\noindent This smoothness constraint enforces a temporal continuity prior on the latent space: temporally adjacent patches that share observational overlap ($S_s < P_s$) should occupy nearby positions in the latent representation. It serves two purposes in the context of data modeling. First, it organizes the latent space along the temporal axis, providing a regularization signal against local observation perturbations; a patch corrupted by sensor noise receives regularization pressure from its temporally adjacent neighbors. Second, it produces latent trajectories whose smoothness properties are directly interpretable: abrupt transitions in the learned trajectory correspond to regime changes in the underlying data, enabling unsupervised pattern discovery.

\subsubsection{Multi-Scale Representation Alignment}

The inter-scale consistency term enforces structural alignment between representations at adjacent temporal scales. For scales $s$ and $s+1$, where the finer scale $s$ has higher temporal resolution ($N_s > N_{s+1}$), it averages the per-pair energy $E_{\text{inter}}$ across all $K-1$ adjacent scale pairs:
\begin{equation}
    \mathcal{L}_{\text{inter}} = \frac{1}{K-1} \sum_{s=1}^{K-1} E_{\text{inter}}(\mathbf{z}^{(s)}, \mathbf{z}^{(s+1)}; \Phi_s)
\end{equation}
where $\text{Pool}: \mathbb{R}^{N_s \times d_l} \to \mathbb{R}^{N_{s+1} \times d_l}$ is a temporal aggregation operator that downsamples the finer scale to the resolution of the coarser scale through windowed averaging, and $\Phi_s: \mathbb{R}^{d_l} \to \mathbb{R}^{d_l}$ is a cross-scale mapping, a two-layer nonlinear transformation that learns the correspondence between latent representations at adjacent scales.

The inter-scale consistency term addresses a structural challenge in multi-view representation learning: when the same underlying data is observed through $K$ different temporal lenses (patch sizes), independent encoding may produce representations that reside in incompatible latent spaces, making cross-scale information fusion unreliable. The consistency term aligns these representations by requiring that a fine-scale representation, when temporally aggregated and transformed, approximates its coarse-scale counterpart. This ensures that all $K$ scales encode mutually consistent representations of the same underlying data, enabling the forecast head to combine multi-scale information without latent space misalignment. Practically, this term also acts as a regularizer that suppresses scale-specific overfitting: features that improve a single scale's fitness but break cross-scale consistency are penalized.

\subsubsection{Regularization Objective and Loss Synergy}

Together, the two regularization terms define a \emph{regularization objective} over the collection of latent representations $\mathcal{Z} = \{\mathbf{z}^{(s)}\}$ and cross-scale mappings $\Phi = \{\Phi_s\}$:
\begin{equation}
    \mathcal{R}(\mathcal{Z}, \Phi) = \lambda_{\text{intra}}\,\mathcal{L}_{\text{intra}} + \lambda_{\text{inter}}\,\mathcal{L}_{\text{inter}}
    \label{eq:regularization}
\end{equation}

The total training objective $\mathcal{L}_{\text{total}} = \mathcal{L}_{\text{MSE}} + \mathcal{R}(\mathcal{Z}, \Phi)$ jointly minimizes forecasting error and structural regularization penalties, with the weights $\lambda_{\text{intra}}, \lambda_{\text{inter}}$ controlling the strength of each penalty. $\mathcal{R}$ is additively separable across scales and scale pairs, allowing each gradient contribution to be computed independently, mirroring the modular architecture of the multi-scale design. These auxiliary losses are designed as standalone regularizers: they can be integrated into any multi-scale architecture that produces per-scale latent representations, and their effectiveness does not depend on the specific choice of backbone. The joint application of two regularization terms ensures that smoothness propagates across the scale hierarchy: fine-scale regularity transmits to coarse scales through cross-scale mapping, while coarse-scale structure constrains the trend direction of fine-scale representations.

\subsection{Scale-Adaptive Forecast Fusion}
\label{sec:forecast_head}

The latent representations from all $K$ scales are fused into the final forecast through a learnable convex combination. Each scale's flattened latent features are independently projected to the prediction horizon via a linear head:
\begin{equation}
    \hat{\mathbf{y}} = \sum_{s=1}^{K} \alpha_s \cdot \mathbf{W}_{\text{out}}^s \cdot \text{flatten}(\mathbf{z}^{(s)}), \quad \alpha_s = \frac{\exp(w_s)}{\sum_{j=1}^{K} \exp(w_j)}
\end{equation}
where $\{w_s\}$ are learnable scale weights, initialized uniformly and normalized through a softmax gate. The per-scale heads are single linear projections of the flattened latent features, so the relative complexity of each scale is controlled exclusively through the learned latent features $\mathbf{z}^{(s)}$ and the fusion weights $\alpha_s$, without additional overhead. Finally, reversible instance normalization is inverted to restore the original data distribution.

\subsection{Computational Complexity}
\label{sec:complexity}

We analyze \model's complexity along the input sequence length $L$. The input is decomposed into $K$ temporal scales, each partitioned into $N_s \approx L / S_s$ patches. The CNN backbone dominates the cost: each depthwise separable block applies a fixed-size kernel over $N_s$ patches, so the backbone costs $\mathcal{O}(N_s)$ per scale and $\mathcal{O}(K \cdot L)$ overall, i.e., $\mathcal{O}(L)$ once the small constant $K$ is fixed. By contrast, PatchTST~\cite{nie2023patchtst} applies pairwise self-attention over the $N_s$ patches at $\mathcal{O}(N_s^2)$ cost, which the depthwise-pointwise factorization avoids, ensuring scalability to long inputs. The embedding, projection, and prediction-head components inherit the $\mathcal{O}(L)$ scaling, and the auxiliary regularization terms introduce negligible overhead; with $K \in \{2, 3\}$, the aggregate time and memory footprint remains a small multiple of the single-scale case.

\section{Experiments}
\label{sec:experiments}

\begin{table*}[htbp!]
    \vspace{-0.1in}
    \caption{Comparison results between \model\ and baselines on ten benchmark datasets in the main comparison. \best{Bold} denotes the best MSE/MAE and \second{underlined} denotes the second-best. The Pearson Correlation Coefficient (PCC) value is listed under each dataset's name. All Illness results among baselines, and TimeMixer results on Exchange, are obtained through experiments; other baseline results are sourced from \cite{wang2024timemixer,wang2025mamba}.}
    \label{tab:effectiveness}
    \renewcommand{\arraystretch}{0.70}
    \centering
    \resizebox{\textwidth}{!}{
        \scriptsize
        \setlength{\tabcolsep}{3pt}
        \setlength{\aboverulesep}{1.5pt}
        \setlength{\belowrulesep}{2.2pt}
        \vspace{1mm}
        \begin{tabular}{c c @{\hspace{6pt}}c @{\hspace{7pt}}c @{\hspace{7pt}}c @{\hspace{7pt}}c @{\hspace{7pt}}c @{\hspace{7pt}}c @{\hspace{7pt}}c @{\hspace{7pt}}c}
            \toprule
            \multicolumn{2}{c}{Models} & \model & \textbf{TimeMixer} & \textbf{iTransformer} & \textbf{PatchTST} & \textbf{Crossformer} & \textbf{TimesNet} & \textbf{DLinear} & \textbf{Autoformer} \\
            \cmidrule(lr){3-3} \cmidrule(lr){4-4} \cmidrule(lr){5-5} \cmidrule(lr){6-6} \cmidrule(lr){7-7} \cmidrule(lr){8-8} \cmidrule(lr){9-9} \cmidrule(lr){10-10}
            \multicolumn{2}{c}{Metric} & MSE/MAE & MSE/MAE & MSE/MAE & MSE/MAE & MSE/MAE & MSE/MAE & MSE/MAE & MSE/MAE \\[1pt]
            \toprule
            \multirow{4}{*}{\rotatebox{90}{ETTm1}} & 96 & \best{0.309}/\best{0.355} & \second{0.320}/\second{0.357} & 0.334/0.368 & 0.329/0.367 & 0.404/0.426 & 0.364/0.387 & 0.338/0.375 & 0.345/0.372 \\ 
            & 192 & \best{0.352}/\best{0.379} & \second{0.361}/\second{0.381} & 0.377/0.391 & 0.367/0.385 & 0.450/0.451 & 0.398/0.404 & 0.374/0.387 & 0.380/0.389 \\ 
            & 336 & \best{0.379}/\best{0.401} & \second{0.390}/\second{0.404} & 0.426/0.420 & 0.399/0.410 & 0.532/0.515 & 0.428/0.425 & 0.410/0.411 & 0.413/0.413 \\ 
            & 720 & \best{0.443}/\best{0.436} & \second{0.454}/0.441 & 0.491/0.459 & \second{0.454}/\second{0.439} & 0.666/0.589 & 0.487/0.461 & 0.478/0.450 & 0.474/0.453 \\ 
            \cmidrule(lr){1-10}
            0.224 & Avg & \best{0.371}/\best{0.393} & \second{0.381}/\second{0.395} & 0.407/0.409 & 0.387/0.400 & 0.513/0.495 & 0.419/0.419 & 0.400/0.406 & 0.403/0.407 \\ 
            \midrule

            \multirow{4}{*}{\rotatebox{90}{ETTm2}} & 96 & \best{0.173}/\best{0.258} & \second{0.175}/\best{0.258} & 0.180/0.264 & \second{0.175}/0.259 & 0.287/0.366 & 0.207/0.305 & 0.187/0.267 & 0.193/0.292 \\ 
            & 192 & \best{0.237}/\best{0.299} & \best{0.237}/\best{0.299} & 0.250/0.309 & 0.241/0.302 & 0.414/0.492 & 0.290/0.364 & 0.249/0.309 & 0.284/0.362 \\ 
            & 336 & \best{0.298}/\best{0.339} & \best{0.298}/\second{0.340} & 0.311/0.348 & 0.305/0.343 & 0.597/0.542 & 0.377/0.422 & 0.321/0.351 & 0.369/0.427 \\ 
            & 720 & \second{0.394}/\best{0.395} & \best{0.391}/\second{0.396} & 0.412/0.407 & 0.402/0.400 & 1.730/1.042 & 0.558/0.524 & 0.408/0.403 & 0.554/0.522 \\ 
            \cmidrule(lr){1-10}
            0.325 & Avg & \best{0.275}/\best{0.323} & \best{0.275}/\best{0.323} & 0.288/0.332 & 0.281/0.326 & 0.757/0.611 & 0.358/0.404 & 0.291/0.333 & 0.350/0.401 \\ 
            \midrule

            \multirow{4}{*}{\rotatebox{90}{ETTh1}} & 96 & \best{0.372}/\best{0.399} & \second{0.375}/\second{0.400} & 0.386/0.405 & 0.414/0.419 & 0.423/0.448 & 0.479/0.464 & 0.384/0.402 & 0.386/\second{0.400} \\ 
            & 192 & \best{0.414}/\second{0.423} & \second{0.429}/\best{0.421} & 0.441/0.436 & 0.460/0.445 & 0.471/0.474 & 0.525/0.492 & 0.436/0.429 & 0.437/0.432 \\ 
            & 336 & \best{0.447}/\best{0.443} & 0.484/\second{0.458} & 0.487/\second{0.458} & 0.501/0.466 & 0.570/0.546 & 0.565/0.515 & 0.491/0.469 & \second{0.481}/0.459 \\ 
            & 720 & \best{0.467}/\best{0.467} & \second{0.498}/\second{0.482} & 0.503/0.491 & 0.500/0.488 & 0.653/0.621 & 0.594/0.558 & 0.521/0.500 & 0.519/0.516 \\ 
            \cmidrule(lr){1-10}
            0.222 & Avg & \best{0.425}/\best{0.433} & \second{0.447}/\second{0.440} & 0.454/0.448 & 0.469/0.455 & 0.529/0.522 & 0.541/0.507 & 0.458/0.450 & 0.456/0.452 \\ 
            \midrule

            \multirow{4}{*}{\rotatebox{90}{ETTh2}} & 96 & \best{0.287}/\best{0.340} & \second{0.289}/\second{0.341} & 0.297/0.349 & 0.302/0.348 & 0.745/0.584 & 0.400/0.440 & 0.340/0.374 & 0.333/0.387 \\ 
            & 192 & \best{0.357}/\best{0.388} & \second{0.372}/\second{0.392} & 0.380/0.400 & 0.388/0.400 & 0.877/0.656 & 0.528/0.509 & 0.402/0.414 & 0.477/0.476 \\ 
            & 336 & \second{0.401}/\second{0.422} & \best{0.386}/\best{0.414} & 0.428/0.432 & 0.426/0.433 & 1.043/0.731 & 0.643/0.571 & 0.452/0.452 & 0.594/0.541 \\ 
            & 720 & \second{0.413}/\best{0.433} & \best{0.412}/\second{0.434} & 0.427/0.445 & 0.431/0.446 & 1.104/0.763 & 0.874/0.679 & 0.462/0.468 & 0.831/0.657 \\ 
            \cmidrule(lr){1-10}
            0.325 & Avg & \best{0.364}/\second{0.396} & \best{0.364}/\best{0.395} & 0.383/0.406 & 0.387/0.407 & 0.942/0.683 & 0.611/0.550 & 0.414/0.427 & 0.559/0.515 \\ 
            \midrule

            \multirow{4}{*}{\rotatebox{90}{ECL}} & 96 & 0.159/0.248 & \second{0.153}/\second{0.247} & \best{0.148}/\best{0.240} & 0.181/0.270 & 0.219/0.314 & 0.237/0.329 & 0.168/0.272 & 0.197/0.282 \\ 
            & 192 & 0.168/0.258 & \second{0.166}/\second{0.256} & \best{0.162}/\best{0.253} & 0.188/0.274 & 0.231/0.322 & 0.236/0.330 & 0.184/0.289 & 0.196/0.285 \\ 
            & 336 & \second{0.184}/\second{0.275} & 0.185/0.277 & \best{0.178}/\best{0.269} & 0.204/0.293 & 0.246/0.337 & 0.249/0.344 & 0.198/0.300 & 0.209/0.301 \\ 
            & 720 & \second{0.222}/\best{0.309} & 0.225/\second{0.310} & 0.225/0.317 & 0.246/0.324 & 0.280/0.363 & 0.284/0.373 & \best{0.220}/0.320 & 0.245/0.333 \\ 
            \cmidrule(lr){1-10}
            0.489 & Avg & 0.183/0.273 & \second{0.182}/\second{0.272} & \best{0.178}/\best{0.270} & 0.205/0.290 & 0.244/0.334 & 0.252/0.344 & 0.193/0.295 & 0.212/0.300 \\ 
            \midrule

            \multirow{4}{*}{\rotatebox{90}{Exchange}} & 96 & \best{0.078}/\best{0.195} & \second{0.085}/\second{0.203} & 0.086/0.206 & 0.088/0.205 & 0.256/0.367 & 0.094/0.218 & 0.107/0.234 & 0.088/0.218 \\ 
            & 192 & \best{0.158}/\best{0.288} & 0.181/0.301 & 0.177/\second{0.299} & \second{0.176}/\second{0.299} & 0.470/0.509 & 0.184/0.307 & 0.226/0.344 & \second{0.176}/0.315 \\ 
            & 336 & \best{0.278}/\best{0.382} & 0.347/0.425 & 0.331/0.417 & \second{0.301}/\second{0.397} & 1.268/0.883 & 0.349/0.431 & 0.367/0.448 & 0.313/0.427 \\ 
            & 720 & \best{0.794}/\best{0.669} & 0.871/0.701 & 0.847/\second{0.691} & 0.901/0.714 & 1.767/1.068 & 0.852/0.698 & 0.964/0.746 & \second{0.839}/0.695 \\ 
            \cmidrule(lr){1-10}
            0.513 & Avg & \best{0.327}/\best{0.384} & 0.371/0.407 & 0.360/\second{0.403} & 0.366/0.404 & 0.940/0.707 & 0.370/0.413 & 0.416/0.443 & \second{0.354}/0.414 \\ 
            \midrule

            \multirow{4}{*}{\rotatebox{90}{Weather}} & 96 & 0.167/\second{0.211} & \second{0.163}/\best{0.209} & 0.174/0.214 & 0.177/0.218 & \best{0.158}/0.230 & 0.202/0.261 & 0.172/0.220 & 0.196/0.255 \\ 
            & 192 & 0.215/\second{0.253} & \second{0.208}/\best{0.250} & 0.221/0.254 & 0.225/0.259 & \best{0.206}/0.277 & 0.242/0.298 & 0.219/0.261 & 0.237/0.296 \\ 
            & 336 & \second{0.271}/\second{0.294} & \best{0.251}/\best{0.287} & 0.278/0.296 & 0.278/0.297 & 0.272/0.335 & 0.287/0.335 & 0.280/0.306 & 0.283/0.335 \\ 
            & 720 & 0.346/\second{0.344} & \best{0.339}/\best{0.341} & 0.358/0.347 & 0.354/0.348 & 0.398/0.418 & 0.351/0.386 & 0.365/0.359 & \second{0.345}/0.381 \\ 
            \cmidrule(lr){1-10}
            0.339 & Avg & \second{0.250}/\second{0.275} & \best{0.240}/\best{0.271} & 0.258/0.278 & 0.259/0.280 & 0.259/0.315 & 0.270/0.320 & 0.259/0.286 & 0.265/0.317 \\ 
            \midrule

            \multirow{4}{*}{\rotatebox{90}{Solar}} & 96 & \second{0.196}/\second{0.243} & \best{0.189}/0.259 & 0.203/\best{0.237} & 0.234/0.286 & 0.310/0.331 & 0.312/0.399 & 0.250/0.292 & 0.290/0.378 \\ 
            & 192 & \second{0.229}/\second{0.264} & \best{0.222}/0.283 & 0.233/\best{0.261} & 0.267/0.310 & 0.734/0.725 & 0.339/0.416 & 0.296/0.318 & 0.320/0.398 \\ 
            & 336 & \second{0.244}/\second{0.278} & \best{0.231}/0.292 & 0.248/\best{0.273} & 0.290/0.315 & 0.750/0.735 & 0.368/0.430 & 0.319/0.330 & 0.353/0.415 \\ 
            & 720 & \second{0.243}/\second{0.277} & \best{0.223}/0.285 & 0.249/\best{0.275} & 0.289/0.317 & 0.769/0.765 & 0.370/0.425 & 0.338/0.337 & 0.356/0.413 \\ 
            \cmidrule(lr){1-10}
            0.916 & Avg & \second{0.228}/\second{0.266} & \best{0.216}/0.280 & 0.233/\best{0.262} & 0.270/0.307 & 0.641/0.639 & 0.347/0.418 & 0.301/0.319 & 0.330/0.401 \\ 
            \midrule

            \multirow{4}{*}{\rotatebox{90}{Traffic}} & 96 & \second{0.437}/\second{0.280} & 0.462/0.285 & \best{0.395}/\best{0.268} & 0.462/0.295 & 0.522/0.290 & 0.805/0.493 & 0.593/0.321 & 0.650/0.396 \\ 
            & 192 & \second{0.443}/\best{0.275} & 0.473/0.296 & \best{0.417}/\second{0.276} & 0.466/0.296 & 0.530/0.293 & 0.756/0.474 & 0.617/0.336 & 0.598/0.370 \\ 
            & 336 & \second{0.451}/\best{0.282} & 0.498/0.296 & \best{0.433}/\second{0.283} & 0.482/0.304 & 0.558/0.305 & 0.762/0.477 & 0.629/0.336 & 0.605/0.373 \\ 
            & 720 & \second{0.487}/\best{0.302} & 0.506/0.313 & \best{0.467}/\best{0.302} & 0.514/0.322 & 0.589/0.328 & 0.719/0.449 & 0.640/0.350 & 0.645/0.394 \\ 
            \cmidrule(lr){1-10}
            0.564 & Avg & \second{0.455}/\second{0.285} & 0.484/0.297 & \best{0.428}/\best{0.282} & 0.481/0.304 & 0.550/0.304 & 0.760/0.473 & 0.620/0.336 & 0.625/0.383 \\ 
            \midrule

            \multirow{4}{*}{\rotatebox{90}{Illness}} & 24 & \best{1.460}/\best{0.733} & 2.034/0.865 & \second{1.530}/\second{0.796} & 2.342/0.982 & 4.395/1.429 & 2.534/1.016 & 2.333/0.918 & 3.421/1.289 \\ 
            & 36 & \best{1.162}/\best{0.674} & 1.806/0.807 & \second{1.498}/\second{0.769} & 2.274/0.968 & 4.096/1.324 & 2.491/0.978 & 1.541/0.777 & 2.859/1.129 \\ 
            & 48 & \best{1.176}/\best{0.720} & 1.891/0.831 & \second{1.456}/\second{0.783} & 2.125/0.935 & 4.044/1.308 & 2.331/0.958 & 1.502/0.787 & 2.633/1.089 \\ 
            & 60 & \best{1.580}/\best{0.808} & 2.131/0.902 & 1.843/0.887 & 2.329/0.999 & 4.296/1.357 & 2.492/1.004 & \second{1.788}/\second{0.875} & 2.937/1.174 \\ 
            \cmidrule(lr){1-10}
            0.708 & Avg & \best{1.345}/\best{0.734} & 1.966/0.851 & \second{1.582}/\second{0.809} & 2.268/0.971 & 4.208/1.355 & 2.462/0.989 & 1.791/0.839 & 2.962/1.170 \\ 
            \bottomrule
        \end{tabular}
    }
    \vspace{-0.1in}
\end{table*}


\textbf{Implementation Details.} We train all models with the Adam optimizer using the OneCycle learning-rate scheduler on RTX4090/4090D GPU servers running PyTorch 2.8.0. Training runs for 15 epochs with a batch size of 32 and early stopping patience of 5, using a fixed random seed of 2023. Hyperparameters are selected by Bayesian search on the validation set. Unless otherwise noted, the input length is $L{=}96$ with forecast horizons $P \in \{96,192,336,720\}$; Illness uses $P \in \{24,36,48,60\}$, and PEMS08 uses $P \in \{12,24,48,96\}$.

\textbf{Datasets.} We evaluate \model\ on ten publicly available benchmark datasets in the main comparison, including ETT (four sub-sets)~\cite{zhou2021informer}, Electricity (ECL), Exchange, Weather, Solar, Traffic, and Illness, and further adopt the traffic benchmark PEMS08 for the ablation study. The eleven datasets span diverse domains, with variable counts ranging from 7 to 862 and temporal granularities from 5~minutes to 1~week. The datasets are split chronologically into training, validation, and test sets at a 70\%/10\%/20\% ratio. Following standard protocol, we adopt Mean Squared Error (MSE) and Mean Absolute Error (MAE) as evaluation metrics.

\textbf{Baselines.} We compare \model\ against seven leading baselines from three architectural families. The first gathers Transformer-based models: \textbf{PatchTST}~\cite{nie2023patchtst} processes channel-independent subseries patches, \textbf{iTransformer}~\cite{liu2023itransformer} inverts attention to operate across variates, \textbf{Crossformer}~\cite{zhang2023crossformer} applies two-stage attention over both time and variables, and \textbf{Autoformer}~\cite{wu2021autoformer} substitutes period-aware auto-correlation for self-attention. The second family covers lightweight models: \textbf{DLinear}~\cite{zeng2023transformers} fits decomposed trend and seasonal components through linear layers, and \textbf{TimeMixer}~\cite{wang2024timemixer} cascades MLPs across multiple temporal resolutions. The third captures a structurally distinct paradigm, where \textbf{TimesNet}~\cite{wu2023timesnet} reshapes one-dimensional series into two-dimensional tensors guided by detected periods.

\subsection{Effectiveness}

Table~\ref{tab:effectiveness} reports results across all ten datasets. \model\ achieves 58 best and 33 second-best placements, exceeding the totals of TimeMixer (26/36) and iTransformer (21/14). The advantage is consistent on benchmarks with weakly correlated channels. Across the strongly periodic ETT datasets, \model\ attains the lowest average MSE outright on ETTm1 and ETTh1 and ties for the best average on ETTm2 and ETTh2. On Exchange, a nearly random-walk financial series, \model\ achieves its strongest relative advantage, outperforming baselines at every horizon. Conversely, on the strongly correlated benchmarks ECL, Traffic, and Solar, channel-dependent models exploit cross-variable interactions to take the lead, with iTransformer leading on ECL and Traffic and TimeMixer on Solar; \model\ nonetheless trails the leader by a narrow margin on all three. On Weather, \model\ ranks second to TimeMixer by a moderate margin, while on Illness, whose 676 training samples and short horizons favor parameter efficiency, it attains the lowest MSE at every horizon. This consistently leading or near-leading performance across heterogeneous benchmarks confirms the effectiveness of the proposed design.

These gains originate in \model's structured latent space, the central innovation of its design. Two complementary penalties, $\mathcal{L}_{\text{intra}}$ and $\mathcal{L}_{\text{inter}}$, organize the latent representation so that all scales encode mutually consistent dynamics, jointly suppressing perturbation propagation and stabilizing long-horizon forecasts. The penalties act on the per-scale representations produced by the remaining components. Multi-scale patching supplies the complementary granularities on which the penalties are defined; the depthwise separable CNN backbone extracts scale-specific features at linear cost; the learnable fusion head adaptively weights each scale to sustain accuracy at both short and long horizons. Together, these mechanisms enable \model\ to match or surpass the baselines in forecasting accuracy, demonstrating the feasibility of combining channel-independent processing with explicit latent-space regularization.

\begin{figure*}[!t]
    \centering
    \includegraphics[width=\linewidth]{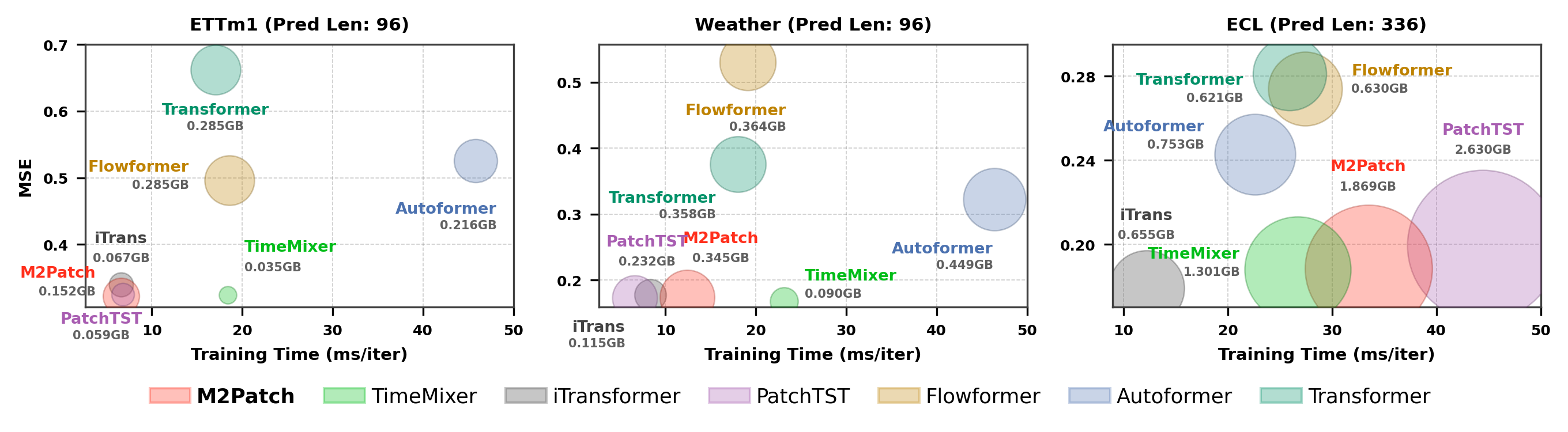}
    \caption{Comparison of training efficiency and forecasting accuracy between \model\ and baseline models on ETTm1, Weather, and ECL.}
    \label{fig:efficiency}
\end{figure*}

\subsection{Efficiency}
\label{sec:efficiency}

Figure~\ref{fig:efficiency} compares \model\ against representative baselines on ETTm1, Weather, and ECL across forecasting accuracy, training time per iteration, and allocated GPU memory. Across all three datasets, \model\ attains a competitive accuracy-efficiency balance. On ETTm1, it records both the shortest training time and the lowest MSE among all evaluated models, at 6.57~ms per iteration versus 18.38~ms for TimeMixer. On Weather, its per-iteration cost of 12.43~ms trails only PatchTST and iTransformer, while its MSE stays within a small margin of the best baselines. On ECL with 321 variate channels, \model\ remains faster than PatchTST and reaches an MSE on par with the leading baselines. GPU memory consumption is also moderate across all three datasets.

These efficiency gains follow directly from \model's architectural design. As analyzed in Section~\ref{sec:complexity}, the multi-scale patch structures and separated projection heads grow the model complexity only by a constant factor, avoiding an exploding memory allocation. Depthwise separable convolution replaces quadratic self-attention with linear-complexity temporal mixing, while the latent projection operates via lightweight MLPs. Because depthwise separable convolution scales linearly in $L$ per scale and the multi-scale decomposition adds only a small constant factor $K$, the per-iteration cost remains modest regardless of dataset width, as reflected in the GPU memory profile shown in Fig.~\ref{fig:efficiency}.

\subsection{Ablation Study}
\label{sec:ablation}

To evaluate the contribution of each architectural component, we conduct a cumulative ablation study on five benchmarks spanning variable counts from 7 to 321. Table~\ref{tab:ablation} reports five variants that progressively remove components relative to the full model, regressing toward a single-scale patch forecaster. Variant~(1), w/o Latent Projection, bypasses the dimensionality-reduction stage so that feature compression through the latent bottleneck no longer applies. Variant~(2), w/o Multi-Scale, collapses the multi-scale patch family to the finest scale, removing cross-granularity interaction. Variant~(3), w/o Temporal Convolution, reduces the CNN kernel to size one, eliminating the dilation hierarchy and leaving position-wise feed-forward processing. Variant~(4), TST Encoder, further replaces the CNN backbone with PatchTST's standard Transformer encoder~\cite{nie2023patchtst}, applying multi-head self-attention over patch indices. The full model and all variants reuse the setting that corresponds to the best trial of a per-(dataset, horizon) hyperparameter search, with each variant being a deterministic rerun under the same search-selected configuration.

\begin{table*}[!htbp]
    \caption{Cumulative component ablation of \model\ on five benchmarks. Each row removes one component relative to the full model, regressing toward a single-scale patch forecaster. \checkmark{} indicates that the component is active and \textendash{} that it is removed. \best{Bold} marks the best result per column. Each row reports the average MSE and MAE over the prediction horizons of each benchmark set.}
    \label{tab:ablation}
    \renewcommand{\arraystretch}{0.85}
    \centering
    \scriptsize
    \resizebox{\textwidth}{!}{
        \setlength{\tabcolsep}{3.5pt}
        \setlength{\aboverulesep}{1pt}
        \setlength{\belowrulesep}{2pt}
        \begin{tabular}{l cccc c c c c c}
            \toprule
            \multirow{2}{*}{Variant}
                 & \multicolumn{4}{c}{\textbf{Components}}
                 & \textbf{PEMS08}
                 & \textbf{Weather}
                 & \textbf{ETTh2}
                 & \textbf{ECL}
                 & \textbf{Illness}
                 \\
            \cmidrule(lr){2-5}\cmidrule(lr){6-6}\cmidrule(lr){7-7}\cmidrule(lr){8-8}\cmidrule(lr){9-9}\cmidrule(lr){10-10}
                 & \makecell{Latent\\Proj.} & \makecell{Multi-\\Scale} & \makecell{Dilated\\Conv} & \makecell{CNN\\Backbone} & MSE/MAE & MSE/MAE & MSE/MAE & MSE/MAE & MSE/MAE \\
            \midrule

                 \makecell[l]{Full \model} & \checkmark{} & \checkmark{} & \checkmark{} & \checkmark{} & \best{0.222}/\best{0.296} & \best{0.250}/\best{0.275} & \best{0.364}/\best{0.396} & \best{0.183}/\best{0.273} & \best{1.345}/\best{0.734} \\ 

            \midrule

                 \makecell[l]{w/o Latent \\ Projection} & \textendash{} & \checkmark{} & \checkmark{} & \checkmark{} & 0.241/0.313 & 0.251/0.276 & 0.382/0.405 & 0.185/0.274 & 1.506/0.780 \\ 

            \midrule

                 \makecell[l]{w/o Multi-Scale} & \textendash{} & \textendash{} & \checkmark{} & \checkmark{} & 0.263/0.337 & 0.252/0.277 & 0.388/0.409 & 0.193/0.285 & 1.555/0.785 \\ 

            \midrule

                 \makecell[l]{w/o Temporal \\ Conv} & \textendash{} & \textendash{} & \textendash{} & \checkmark{} & 0.321/0.373 & 0.260/0.281 & 0.395/0.416 & 0.210/0.301 & 1.938/0.876 \\ 

            \midrule

                 \makecell[l]{TST Encoder} & \textendash{} & \textendash{} & \textendash{} & \textendash{} & 0.254/0.331 & 0.266/0.290 & 0.400/0.421 & 0.196/0.289 & 1.769/0.837 \\ 
            \bottomrule
        \end{tabular}
    }
\end{table*}


The two variants that preserve the CNN backbone expose a consistent hierarchy of component contributions. Bypassing the latent projection leaves average MSE nearly unchanged on Weather, no more than 5\% higher on ECL and ETTh2, but raises it by 8.6\% on PEMS08 and 12.0\% on Illness. The compression bottleneck thus acts as a structural regularizer whose absence is almost harmless when supervision is plentiful and becomes most damaging where the model must generalize from limited evidence, namely the small-sample Illness benchmark and the long-range traffic regime. Collapsing the multi-scale patch family to the finest scale deepens the same contrast: degradation stays below 1\% on Weather and reaches 5.5\% to 18.5\% on the remaining benchmarks, with the largest losses of 15.6\% and 18.5\% on Illness and PEMS08. Cross-granularity redundancy therefore pays off as channel coupling strengthens and forecast ranges lengthen, since coarse scales anchor the trend that fine-scale patches alone cannot extrapolate reliably.

Reducing the CNN kernel to size one is by far the most destructive step, confirming that dilated temporal convolution is the core temporal-mixing mechanism of the backbone. Average MSE rises by 44.6\% on PEMS08 and 44.1\% on Illness, by 14.8\% on ECL, and by at most 8.5\% on the remaining benchmarks. The exponentially growing dilation constructs a hierarchical receptive field matched to the forecast range, so its absence is felt most where long-range dependence and cross-channel coupling prevail, whereas on strongly periodic series position-wise processing still recovers much of the accuracy. Replacing the CNN backbone with a Transformer encoder partially reverses this damage: on PEMS08, ECL, and Illness, global self-attention restores cross-patch interaction and recovers part of the error incurred by the kernel-one configuration, yet it trails the full model on all five benchmarks by 6.4\% to 31.5\% in average MSE and even underperforms the kernel-one CNN on Weather and ETTh2. The dilation hierarchy therefore supplies an inductive bias that global attention does not replicate, and the gap widens as supervision grows scarcer. Overall, the ablation attributes the largest share of the accuracy gain to the dilated CNN backbone, while multi-scale patching and the latent projection add complementary contributions that grow as sample size shrinks and channel coupling strengthens.

\begin{table}[!htbp]
    \caption{Ablation results for auxiliary regularization terms on four datasets. The Baseline uses a squared first-order difference penalty (TV1) for intra-scale smoothness and MSE with a learnable cross-scale mapping for inter-scale consistency. (a)~Substitution replaces each loss term with an alternative form at equal weight: Intra-Var maximizes variance instead of enforcing smoothness, and Inter-NoMap removes the learnable cross-scale mapping. (b)~Ablation sets one or both regularization weights to zero. \best{Bold} marks the best in the three-way substitution comparison, while \lossmark{underlining} marks the best in the ablation comparison; Baseline cells may carry both marks.}
    \label{tab:ablation_loss}
    \renewcommand{\arraystretch}{0.80}
    \renewcommand{\baselinestretch}{1.1}\selectfont
    \centering
    \resizebox{0.7\textwidth}{!}{
        \setlength{\tabcolsep}{6pt}
        \setlength{\aboverulesep}{1pt}
        \setlength{\belowrulesep}{2pt}
        \begin{tabular}{l c c c c c}
            \toprule
            \rule{0pt}{4pt}
            \multirow{2}{*}{Variant} & \multirow{2}{*}{Len} & \textbf{ETTh1} & \textbf{Weather} & \textbf{ECL} & \textbf{Illness} \\
            \cmidrule(lr){3-6}
            & & \rule{0pt}{4pt} MSE/MAE & MSE/MAE & MSE/MAE & MSE/MAE \\
            \midrule

            \multirow{5}{*}{\parbox{1.5cm}{Baseline}}
            & 96 & \lossmark{0.375}/\lossmark{0.401} & 0.170/\best{0.213} & \best{0.161}/\best{0.251} & 2.273/0.932 \\ 
            & 192 & \best{0.420}/\best{0.429} & \best{\lossmark{0.216}}/\best{\lossmark{0.253}} & \best{0.172}/\best{0.261} & 2.188/\best{0.906} \\ 
            & 336 & \best{\lossmark{0.450}}/\best{\lossmark{0.444}} & \best{\lossmark{0.271}}/\best{\lossmark{0.294}} & \best{0.189}/\best{0.279} & 1.904/0.866 \\ 
            & 720 & \best{\lossmark{0.471}}/\best{\lossmark{0.471}} & \best{\lossmark{0.346}}/\best{\lossmark{0.344}} & \best{0.228}/\best{0.312} & 2.415/1.067 \\ 
            \cmidrule(r){2-6}
            & Avg & \best{\lossmark{0.429}}/\best{\lossmark{0.436}} & \best{\lossmark{0.251}}/\best{\lossmark{0.276}} & \best{0.188}/\best{0.276} & 2.195/0.943 \\ 
            \midrule

            \multicolumn{6}{l}{\textit{(a) Substitution: loss form replacement}} \\
            \cmidrule(lr){1-6}

            \multirow{5}{*}{\parbox{1.5cm}{Intra-Var}}
            & 96 & 0.423/0.423 & 0.205/0.248 & 0.227/0.316 & \best{2.156}/\best{0.914} \\ 
            & 192 & 0.464/0.445 & 0.248/0.280 & 0.248/0.338 & \best{2.166}/0.914 \\ 
            & 336 & 0.493/0.457 & 0.295/0.310 & 0.252/0.343 & \best{1.720}/\best{0.816} \\ 
            & 720 & 0.487/0.478 & 0.368/0.357 & 0.310/0.384 & \best{2.139}/\best{0.931} \\ 
            \cmidrule(r){2-6}
            & Avg & 0.467/0.451 & 0.279/0.299 & 0.259/0.345 & \best{2.045}/\best{0.894} \\ 

            \multirow{5}{*}{\parbox{1.5cm}{Inter-NoMap}}
            & 96 & \best{0.373}/\best{0.399} & \best{0.169}/0.214 & 0.162/0.252 & 2.378/0.963 \\ 
            & 192 & 0.432/0.433 & \best{0.216}/0.255 & 0.173/0.262 & 2.561/1.008 \\ 
            & 336 & 0.452/0.445 & 0.272/\best{0.294} & 0.190/\best{0.279} & 2.166/0.919 \\ 
            & 720 & 0.481/0.473 & 0.347/\best{0.344} & 0.231/0.315 & 2.413/0.992 \\ 
            \cmidrule(r){2-6}
            & Avg & 0.435/0.438 & \best{0.251}/0.277 & 0.189/0.277 & 2.380/0.971 \\ 
            \midrule

            \multicolumn{6}{l}{\textit{(b) Ablation: term removal}} \\
            \cmidrule(lr){1-6}

            \multirow{5}{*}{\parbox{1.5cm}{w/o $\mathcal{L}_{\text{intra}}$}}
            & 96 & 0.387/0.403 & 0.173/0.216 & 0.161/0.251 & 2.178/0.892 \\ 
            & 192 & \lossmark{0.415}/\lossmark{0.423} & 0.218/0.254 & 0.171/0.261 & 1.678/0.783 \\ 
            & 336 & 0.471/0.455 & 0.275/0.296 & 0.190/0.282 & 1.840/0.862 \\ 
            & 720 & 0.485/0.472 & 0.351/0.347 & 0.226/0.311 & 1.969/0.899 \\ 
            \cmidrule(r){2-6}
            & Avg & 0.440/0.438 & 0.254/0.278 & 0.187/0.276 & 1.916/0.859 \\ 

            \multirow{5}{*}{\parbox{1.5cm}{w/o $\mathcal{L}_{\text{inter}}$}}
            & 96 & 0.378/0.402 & \lossmark{0.169}/\lossmark{0.212} & 0.161/0.251 & 2.181/0.890 \\ 
            & 192 & 0.417/0.428 & 0.218/0.255 & 0.172/0.261 & 1.917/0.840 \\ 
            & 336 & 0.453/\lossmark{0.444} & 0.272/0.295 & 0.190/0.281 & 1.638/0.805 \\ 
            & 720 & 0.488/0.476 & 0.350/0.346 & 0.226/0.311 & 2.304/1.030 \\ 
            \cmidrule(r){2-6}
            & Avg & 0.434/0.438 & 0.252/0.277 & 0.187/0.276 & 2.010/0.891 \\ 

            \multirow{5}{*}{\parbox{1.5cm}{Pure Forecast}}
            & 96 & 0.393/0.408 & 0.173/0.215 & \lossmark{0.159}/\lossmark{0.248} & \lossmark{1.460}/\lossmark{0.733} \\ 
            & 192 & 0.440/0.441 & 0.218/0.255 & \lossmark{0.168}/\lossmark{0.258} & \lossmark{1.162}/\lossmark{0.674} \\ 
            & 336 & 0.473/0.449 & 0.274/0.295 & \lossmark{0.184}/\lossmark{0.275} & \lossmark{1.176}/\lossmark{0.720} \\ 
            & 720 & 0.524/0.499 & 0.351/0.347 & \lossmark{0.222}/\lossmark{0.309} & \lossmark{1.580}/\lossmark{0.808} \\ 
            \cmidrule(r){2-6}
            & Avg & 0.458/0.449 & 0.254/0.278 & \lossmark{0.183}/\lossmark{0.273} & \lossmark{1.345}/\lossmark{0.734} \\ 

            \bottomrule
        \end{tabular}
    }
\end{table}


\subsection{Analysis of Regularization Terms}
\label{sec:analysis_loss}

Table~\ref{tab:ablation_loss} isolates the contribution of each regularization term through two complementary comparisons that share a common Baseline: an intra-scale smoothness penalty and an MSE inter-scale consistency term, both evaluated under a unified weight $\lambda$. Panel~(a) substitutes each loss with an alternative mathematical form at equal weight, testing whether the chosen formulation is preferable to plausible alternatives; Panel~(b) zeroes one or both regularization weights, testing whether regularization is necessary at all. The four datasets cover channel counts from 7 to 321, spanning both low- and high-dimensional regimes.

\vpara{Regularization Necessity.} On ETTh1 and Weather, the Baseline is simultaneously optimal in both panels, confirming that the full regularization is beneficial: removing both terms degrades average MSE by 6.8\% on ETTh1, and the gap widens at longer horizons. The outcome inverts on the high-dimensional ECL and the small-sample Illness datasets, where Pure Forecast is optimal. The penalty is modest on ECL but severe on Illness, where the average MSE drops sharply, reflecting how extremely scarce training samples cause regularization to over-constrain the latent representation.

\vpara{Loss Form Substitution.} Replacing the smoothness penalty with a max-variance objective is harmful on high-dimensional benchmarks: average MSE rises by 37.8\% on ECL, with consistent degradation across every horizon, because maximizing dispersion destroys the temporal coherence that structured representations require when variables are numerous. On Illness, however, the same substitution reduces MSE, as anti-collapse dispersion counteracts representation degeneration under scarce supervision. Removing the learnable cross-scale mapping is nearly harmless on data-rich regimes but degrades Illness substantially, indicating that adaptive cross-scale alignment becomes critical for sparse data. On ECL, Pure Forecast is optimal in the necessity comparison yet the Baseline wins the form comparison, revealing that whether and how to regularize are distinct design decisions. The learnable smoothness penalty directly models the intrinsic temporal dynamics among adjacent patches; when training data are sufficient, it outperforms the dispersion-seeking alternative on most datasets, and only under severe sample scarcity does the latter's anti-collapse effect become preferable.

Beyond accuracy, these penalties underpin the robustness and representation analyses of Sections~\ref{sec:robustness} and~\ref{sec:clustering}: the same smoothness and alignment terms that yield the gains above dampen burst-missing corruption and organize channels by physical function, indicating that the regularization improves the representation itself rather than merely tuning the prediction head.

Together, these results establish the smoothness penalty combined with learnable cross-scale mapping as a robust solution: it avoids severe form-level degradation across all tested datasets, and its degradation relative to the best per-dataset form choice stays within 9\%. The Illness reversal, where anti-collapse and regularization-free configurations excel, delineates a boundary condition for small-sample regimes and motivates adaptive loss selection as future work.

\begin{table}[!htbp]
    \caption{Analysis results for multi-scale configuration, where 3-Scale, 2-Scale, and 1-Scale denote configurations with $K \in \{3,2,1\}$ temporal granularities respectively, to quantify the benefit of multi-granularity temporal modeling.}
    \label{tab:ablation_scale}
    \renewcommand{\arraystretch}{0.80}
    \renewcommand{\baselinestretch}{1.1}\selectfont
    \centering
    \resizebox{0.7\textwidth}{!}{
        \setlength{\tabcolsep}{6pt}
        \setlength{\aboverulesep}{1pt}
        \setlength{\belowrulesep}{2pt}
        \begin{tabular}{l c c c c c}
            \toprule
            \rule{0pt}{4pt}
            \multirow{2}{*}{Variant}
                 & \multirow{2}{*}{Len}
                 & \textbf{ETTh1}
                 & \textbf{Weather}
                 & \textbf{ECL}
                 & \textbf{Illness} \\
            \cmidrule(lr){3-6}
                 &                                      & \rule{0pt}{4pt} MSE/MAE & MSE/MAE & MSE/MAE & MSE/MAE \\
            \midrule

            \multirow{5}{*}{\parbox{1.5cm}{3-Scale}}
                 & 96 & 0.375/0.401 & 0.170/0.213 & \best{0.159}/\best{0.248} & \best{1.460}/\best{0.733} \\ 
                 & 192 & \best{0.414}/\best{0.423} & 0.219/0.257 & \best{0.168}/\best{0.258} & \best{1.162}/\best{0.674} \\ 
                 & 336 & 0.454/0.447 & \best{0.271}/\best{0.294} & \best{0.184}/\best{0.275} & \best{1.176}/\best{0.720} \\ 
                 & 720 & \best{0.465}/\best{0.467} & \best{0.346}/\best{0.344} & \best{0.222}/\best{0.309} & \best{1.580}/\best{0.808} \\ 
            \cmidrule(r){2-6}
                 & Avg & 0.427/0.435 & 0.252/0.277 & \best{0.183}/\best{0.273} & \best{1.345}/\best{0.734} \\ 
            \midrule

            \multirow{5}{*}{\parbox{1.5cm}{2-Scale}}
                 & 96 & \best{0.372}/\best{0.399} & \best{0.167}/\best{0.211} & 0.160/0.251 & 1.705/0.772 \\ 
                 & 192 & 0.417/0.424 & \best{0.215}/\best{0.253} & 0.169/0.260 & 1.285/0.703 \\ 
                 & 336 & \best{0.447}/\best{0.443} & 0.272/0.295 & 0.188/0.278 & 1.548/0.827 \\ 
                 & 720 & 0.467/\best{0.467} & 0.349/0.345 & 0.224/0.311 & 1.664/0.839 \\ 
            \cmidrule(r){2-6}
                 & Avg & \best{0.426}/\best{0.433} & \best{0.251}/\best{0.276} & 0.185/0.275 & 1.551/0.785 \\ 
            \midrule

            \multirow{5}{*}{\parbox{1.5cm}{1-Scale}}
                 & 96 & 0.375/0.401 & 0.170/0.213 & 0.165/0.262 & 1.920/0.871 \\ 
                 & 192 & 0.452/0.442 & 0.216/0.254 & 0.174/0.270 & 1.341/0.721 \\ 
                 & 336 & 0.457/0.449 & \best{0.271}/\best{0.294} & 0.194/0.289 & 1.549/0.806 \\ 
                 & 720 & 0.491/0.481 & \best{0.346}/\best{0.344} & 0.232/0.321 & 1.807/0.870 \\ 
            \cmidrule(r){2-6}
                 & Avg & 0.444/0.443 & \best{0.251}/\best{0.276} & 0.191/0.286 & 1.654/0.817 \\ 

            \bottomrule
        \end{tabular}
    }
\end{table}


\subsection{Analysis of Multi-Scale Configuration}
\label{sec:analysis_scale}

As shown in Table~\ref{tab:ablation_scale}, multi-scale decomposition is beneficial on most datasets: 3-Scale attains the best average MSE on ECL and Illness, 2-Scale is marginally ahead on ETTh1 (0.426 vs 0.427), and the configurations are nearly equivalent on Weather (0.251--0.252). The fine scale acts as a local dynamics encoder whose aggregated information, propagated through cross-scale mapping, stabilizes coarse-scale trend extrapolation. Degradation of the 1-Scale variant is most severe on Illness, where scarce training samples amplify the loss of cross-scale redundancy, and nearly absent on Weather, where slow low-frequency trends dominate and the additional scales act mainly as regularizers; this contrast indicates that multi-scale decomposition is most valuable for data with rich periodic structure.

\begin{figure*}[!t]
    \centering
    \includegraphics[width=\linewidth]{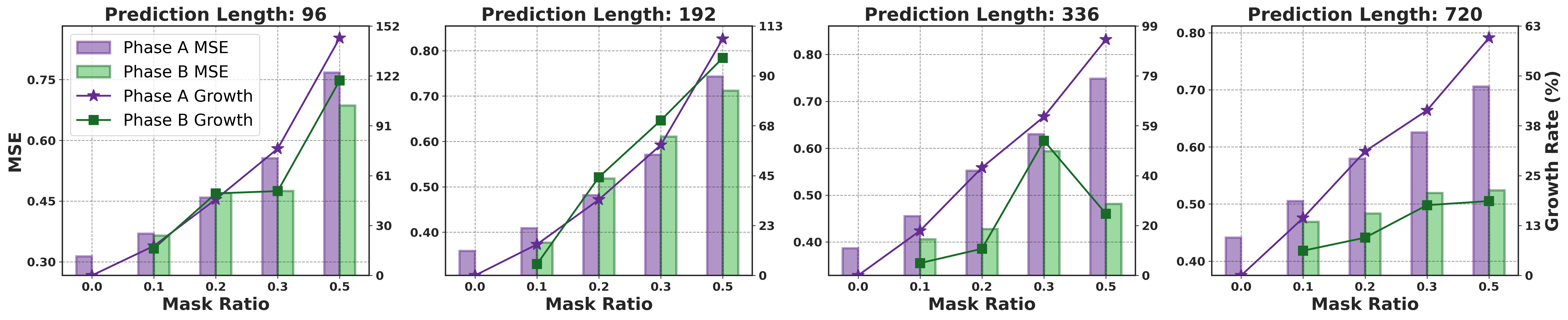}
    \caption{Patch-level missing-value robustness of \model\ on ETTm1 across four forecast horizons. Bars report absolute test MSE under Phase~A and Phase~B; curves report the relative MSE increase over the clean baseline.}
    \label{fig:missing}
\end{figure*}

\subsection{Robustness}
\label{sec:robustness}

We evaluate \model's robustness through a patch-level missing-value experiment that simulates burst missing, a deployment failure mode in which entire contiguous observation blocks are lost to sensor outages rather than corrupted by independent per-step perturbations.

To assess resilience against such corruption, we inject missing values by zeroing randomly selected patches at the finest patching scale with mask ratio $r \in \{0.1, 0.2, 0.3,$ $ 0.5\}$, applied to all channels simultaneously. Phase~A trains the model on clean sequences and tests it on masked inputs, quantifying the inherent robustness of latent representations that have never encountered missing patterns. Phase~B applies the same mask ratio during both training and testing, revealing how much capacity the model recovers once it is allowed to adapt to burst-missing structure. Both phases are normalized against the shared Phase~A clean baseline.

Figure~\ref{fig:missing} reports absolute MSE and relative growth for the two phases. Phase~A exposes an intrinsic robustness that strengthens with forecast horizon: under heavy masking, the relative MSE increase is most severe at the shortest horizon and attenuates considerably at longer horizons, as long-horizon forecasts rely on coarse-scale trend representations that survive fine-scale corruption. Phase~B recovers most of the loss, with the recovery widening as the horizon grows. With missing-aware training, degradation is substantially contained across all horizons, and under light masking the longer horizons incur only marginal loss relative to the clean baseline. These properties stem from the two auxiliary constraints. The temporal continuity term propagates smoothness from adjacent uncorrupted patches, damping the impact of isolated missing blocks. The cross-scale alignment term enables partial reconstruction of missing fine-scale segments from their intact coarse-scale counterparts. Together they prevent partial input corruption from severely degrading the latent encoding, confirming that the structured latent space captures the data's intrinsic dynamics rather than memorizing the full observation.

\begin{figure*}[!t]
    \centering
    \includegraphics[width=\linewidth]{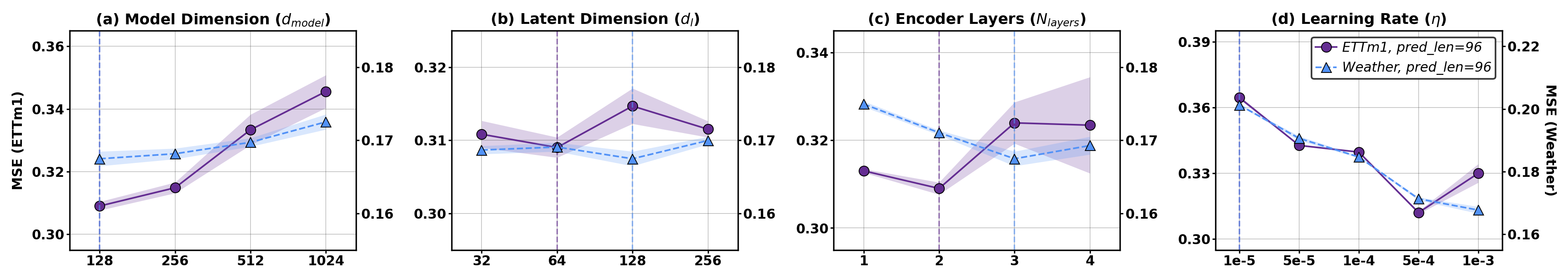}
    \caption{Hyperparameter sensitivity of \model\ on ETTm1 and Weather at the 96-step horizon. Each panel sweeps one hyperparameter: (a)~model dimension, (b)~latent dimension, (c)~encoder layers, (d)~learning rate. Solid lines and shaded bands report the minimum MSE across five seeds with one standard deviation; dashed vertical lines mark the nearest swept value to the per-dataset search-selected setting.}
    \label{fig:sensitivity}
\end{figure*}

\subsection{Hyperparameter Sensitivity Analysis}
\label{sec:sensitivity}

We analyze the sensitivity of \model's forecasting accuracy to its core hyperparameters through a single-factor perturbation protocol on ETTm1 and Weather at the 96-step horizon. Each run varies exactly one hyperparameter while the remaining settings stay at the per-dataset configuration selected by the Bayesian search, and every configuration is trained with five random seeds so that both the best achievable accuracy and its run-to-run variation can be assessed. Figure~\ref{fig:sensitivity} summarizes the results.

The learning rate is the most influential hyperparameter: on ETTm1 the minimum MSE first decreases and then rises, attaining its optimum around $5\times10^{-4}$, whereas on Weather it declines monotonically toward $10^{-3}$, indicating that both datasets attain their optimum within $5\times10^{-4}$ to $10^{-3}$ and suffer pronounced under-fitting below $10^{-4}$. Increasing the model dimension beyond 128 yields no accuracy gain on either dataset, and the latent dimension is the least sensitive setting, whose minimum MSE varies by less than 3\% across the entire swept range on both datasets. Encoder depth exhibits dataset-dependent optima: ETTm1 performs best with two layers and degrades markedly with four, where the cross-seed standard deviation also widens, whereas Weather favors three layers.

These patterns are consistent with the roles of the corresponding design choices. The learning rate governs convergence quality under the OneCycle scheduler, so overly small rates leave the model under-fit, and its narrow effective range motivates the per-dataset search. The redundant capacity introduced by oversized or overdeep encoders provides no accuracy benefit and is most harmful on ETTm1, whose strong periodicity can already be captured by a compact model, whereas the wider Weather benchmark tolerates and even benefits from greater depth; the widening seed variance of deep configurations further indicates that excess capacity destabilizes training. The insensitivity to the latent dimension confirms that the structured latent compression remains effective as long as a bottleneck is present, without requiring precise dimensional tuning. Overall, only the learning rate demands careful selection, while the architectural hyperparameters are robust across wide ranges, a property that favors deployment.

\begin{figure}[!t]
    \centering
    \includegraphics[width=\textwidth]{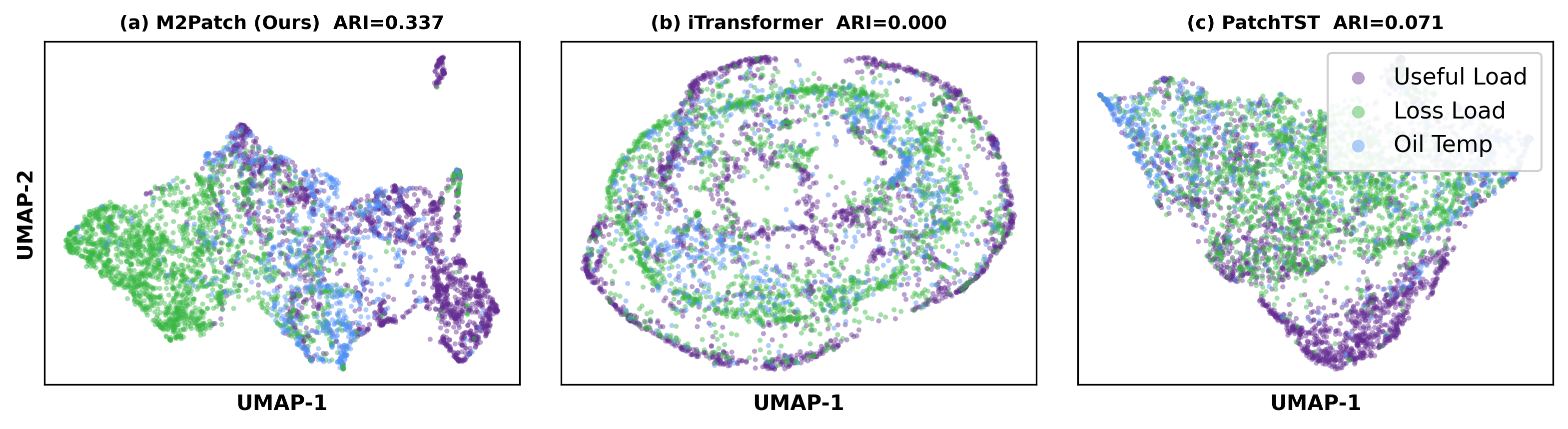}
    \caption{Two-dimensional projection of frozen-encoder window-channel embeddings on ETTm1, with three ground-truth functional groups (Useful Load, Loss Load, Oil Temperature) marked by color. (a)~\model. (b)~iTransformer and (c)~PatchTST under the identical protocol.}
    \label{fig:clustering}
\end{figure}

\subsection{Channel-Relationship Structure}
\label{sec:clustering}

Beyond forecasting accuracy, we ask whether the learned latent space carries channel-relationship structure. We freeze the \model\ checkpoint trained on ETTm1 and extract per-channel embeddings for each test window by concatenating the per-scale latent projections and averaging across patches. The seven transformer channels admit an unambiguous functional partition into useful-load currents, loss-load currents, and the thermally driven oil-temperature signal, a grouping that is never disclosed during training. Under an identical protocol, we cluster these embeddings with KMeans~\cite{macqueen1967kmeans} and project them via UMAP~\cite{mcinnes2018umap}, applying the same steps to frozen iTransformer~\cite{liu2023itransformer} and PatchTST~\cite{nie2023patchtst} encoders. Figure~\ref{fig:clustering} shows that \model\ organizes the three groups into partially separated regions, with an Adjusted Rand Index~\cite{hubert1985ari} of 0.337, whereas iTransformer and PatchTST collapse the channels into a nearly unstructured mixture with ARI values of 0.071 and close to zero, respectively. Since clustering the raw windowed series already attains an ARI of 0.262, \model\ preserves and sharpens the partial functional signal carried by the raw amplitude, while the Transformer backbones destroy it. The forecasting-only objective therefore does not induce channel-relationship-aware representations; the structural priors embedded in \model\ are what uncover the underlying physical grouping.

The contrast stems from how each architecture organizes temporal content. iTransformer compresses the per-channel sequence into a single token, discarding the local waveform shapes that distinguish useful-load from loss-load currents. PatchTST shares the channel-independent patching paradigm but operates at a single fixed granularity without cross-scale alignment or latent regularization, so its near-zero ARI shows that patching alone does not recover functional grouping without explicit structural constraints. \model\ instead preserves this waveform morphology through multi-scale overlapping patches, and the two auxiliary losses draw channels governed by similar dynamics together: $\mathcal{L}_{\text{intra}}$ enforces smooth transitions between adjacent patches, while $\mathcal{L}_{\text{inter}}$ aligns representations across granularities. All three models apply reversible instance normalization~\cite{kim2022reversible}, and applying the same normalization to raw windows without learning yields a near-zero ARI, isolating the contribution of the structured latent constraints from amplitude and normalization effects. This positions \model\ as a forecaster whose latent representation, beyond predicting accurately, captures channel-relationship structure that may benefit downstream tasks such as unsupervised channel profiling or anomaly attribution.

\begin{figure*}[!t]
    \centering
    \includegraphics[width=\linewidth]{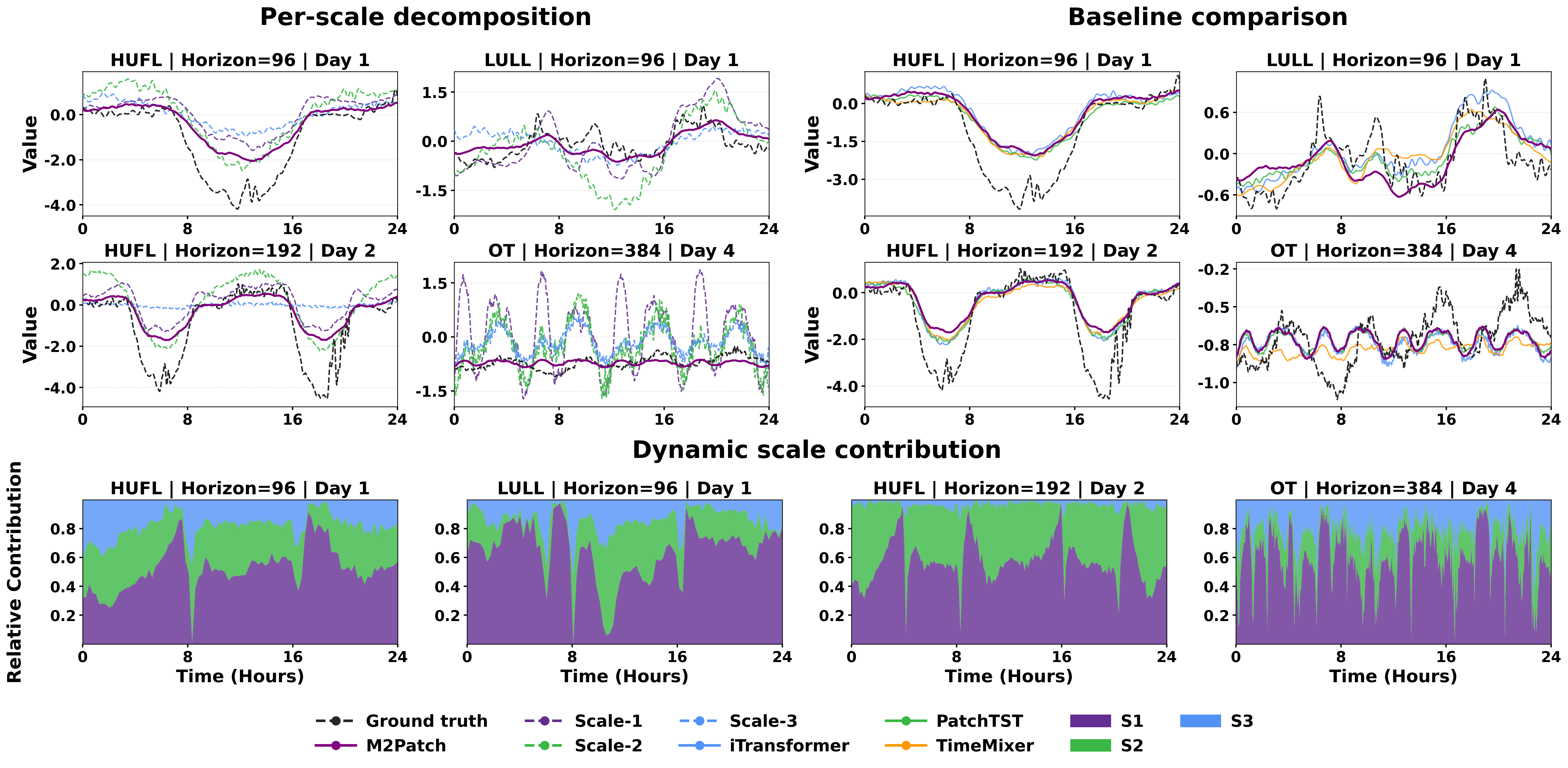}
    \caption{Case study on ETTm1 across four scenarios. Top-left: comparison of \model's fused prediction against individual per-scale predictions. Top-right: baseline comparison of \model\ against iTransformer, PatchTST, and TimeMixer. Bottom: dynamic per-scale effective contribution over the prediction horizon (S1: fine, S2: medium, S3: coarse).}
    \label{fig:case-study}
\end{figure*}

\subsection{Case Study}

Figure~\ref{fig:case-study} presents a multivariate case study on ETTm1 across four representative scenarios. The top-left panel compares \model's fused prediction against each scale's independent prediction; the top-right panel contrasts \model~against best baselines under identical conditions; the bottom row traces the per-scale effective contribution along the prediction horizon.

\vpara{Multi-Scale Decomposition.} Across all four scenarios, the fused prediction tracks the ground truth more closely than any single-scale prediction: fine-scale predictions capture rapid local fluctuations but carry high-frequency noise, coarse-scale predictions follow the trend envelope but miss transients, and the fused output synthesizes both.

\vpara{Baseline Comparison.} \model\ outperforms or matches the three baselines across all four scenarios. On OT at $P{=}384$, where thermal inertia produces smooth multi-day dynamics, \model's coarse-scale contributions dominate and the fused prediction aligns closely with the ground truth, while PatchTST overshoots and TimeMixer lags the turning points. On HUFL at $P{=}96$, where rapid load fluctuations dominate, \model's fine-scale patching captures transient peaks that iTransformer and PatchTST miss. These qualitative observations are consistent with the quantitative results in Table~\ref{tab:effectiveness}.

\vpara{Dynamic Scale Contribution.} The per-scale \emph{effective} contribution, defined as $|w_s \cdot \hat{y}_s(t)| / \sum_{s'} |w_{s'} \cdot \hat{y}_{s'}(t)|$, evolves along the horizon even though the fusion weights $\{w_s\}$ are fixed after training. At the onset of the horizon, the fine-grained scale takes the lead and supplies the short-term detail that near-step prediction demands; as trend-dominant events unfold, the coarse scale assumes the leading role. The OT variable exhibits a more intricate rhythm, with coarse and fine scales alternating in dominance, reconciling short-term volatility with slower trend evolution. This dynamic reweighting corroborates the design principle that multi-scale decomposition furnishes a redundant yet complementary representation from which the forecast head recruits the most informative granularity at every prediction step.

\section{Conclusion}
\label{sec:conclusion}

In this paper, we propose \model, which organizes channel-independent multivariate time series into a structured multi-scale latent space through multi-scale patching and two complementary differentiable penalties. The intra-scale smoothness and inter-scale alignment terms jointly enforce two structural properties at the core of \model's latent organization, converting raw observations into a compact structured representation in which all scales encode mutually consistent dynamics. The depthwise separable CNN backbone extracts scale-specific features at linear complexity in the input length. Experiments on ten benchmarks show that \model\ matches or surpasses the best-performing baselines across the evaluated settings, with the most decisive gains on the weakly correlated ETT benchmarks and pronounced relative advantages on Exchange and Illness. Ablation studies show the complementary contributions of all components: the CNN backbone, the latent projection, and multi-scale patching each sustain accuracy across datasets of varying intervariate correlations, while the two regularization terms further improve both accuracy and robustness through latent structure modeling.

\noindent\textbf{Limitations and Future Work.} The channel-independent design, while robust to distribution shift, cannot exploit explicit cross-variable interactions, so channel-dependent baselines retain a narrow edge on strongly coupled benchmarks such as ECL and Traffic, yet \model\ remains a close competitor. The multi-scale configuration is specified manually from dominant periods, and the fixed-weight regularization can over-constrain the latent space under severe sample scarcity. Future work will pursue data-driven scale discovery, cross-variable causal inference over the learned latent trajectories, and downstream evaluation of the regularized representation on classification and anomaly detection.

\bibliographystyle{elsarticle-num}
\bibliography{ref}

\end{document}